\documentclass[10pt,twocolumn,letterpaper]{article}

\usepackage{iccv}
\usepackage{times}
\usepackage{epsfig}
\usepackage{graphicx}
\usepackage{amsmath}
\usepackage{amssymb}
\usepackage{mmstyle}
\usepackage{mycmd}

\usepackage[breaklinks=true,colorlinks,bookmarks=false]{hyperref}

\iccvfinalcopy 

\def\iccvPaperID{****} 

\begin{document}
	
	\title{\vspace{-0.55cm}A Graph-Based Framework to Bridge Movies and Synopses\vspace{-0.35cm}}
	\author{Yu Xiong$^{1}$ \quad Qingqiu Huang$^{1}$ \quad Lingfeng Guo$^2$ \quad Hang Zhou$^1$ \quad Bolei Zhou$^1$ \quad Dahua Lin$^1$\\
		$^1$CUHK - SenseTime Joint Lab, The Chinese University of Hong Kong \\
		$^2$University of California, Berkeley\\
		{\tt\small \{xy017,hq016,bzhou,dhlin\}@ie.cuhk.edu.hk}\hspace{10pt}
		{\tt\small zhouhang@link.cuhk.edu.hk}\hspace{10pt}
		{\tt\small lingfeng\_guo@berkeley.edu}
	}
	
	\maketitle



\begin{abstract}
\vspace{-10px}
Inspired by the remarkable advances in video analytics,
research teams are stepping towards a greater ambition
-- movie understanding.
However, compared to those activity videos in conventional datasets,
movies are significantly different.
Generally, movies are much longer and consist of much richer temporal
structures. More importantly, the interactions among characters play
a central role in expressing the underlying story.
To facilitate the efforts along this direction, we construct a dataset
called \emph{Movie Synopses Associations (MSA)} over $327$ movies,
which provides a synopsis for each movie, together with annotated associations
between synopsis paragraphs and movie segments.
On top of this dataset, we develop a framework to perform matching between
movie segments and synopsis paragraphs. This framework integrates different
aspects of a movie, including event dynamics and character interactions,
and allows them to be matched with parsed paragraphs, based on a graph-based
formulation.
Our study shows that the proposed framework remarkably improves the matching
accuracy over conventional feature-based methods.
It also reveals the importance of narrative structures and character
interactions in movie understanding.
Dataset and code are available at: \url{https://ycxioooong.github.io/projects/moviesyn}
\end{abstract}


\begin{figure*}[t]
	\centering
	\includegraphics[width=0.9\linewidth]{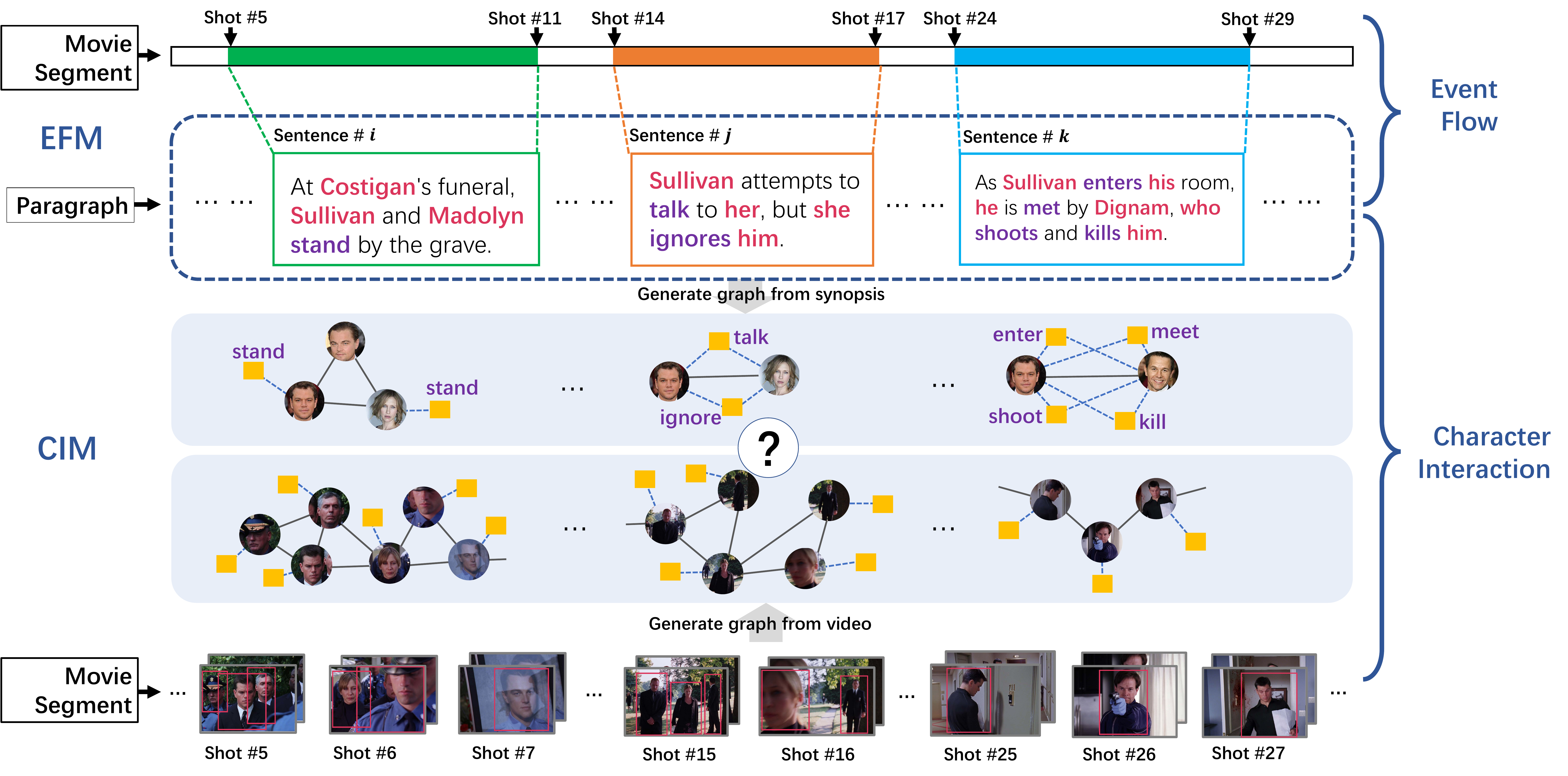}
	\caption{\small
	The story in a synopsis paragraph is
	presented following narrative structures (the upper part),
	which are modeled into \emph{Event Flow Module};
	The lower part shows the character interaction captured in \emph{Character Interaction Module}. 
	The yellow squares denote action.
	}
	\label{fig:teaser}
	\vspace{-10pt}
\end{figure*}

\vspace{-10pt}
\section{Introduction}
\label{sec:intro}
\vspace{-5pt}
%
Among various forms of media, movies are often considered as the best to
convey stories. While creating a movie, the director can leverage a variety
of elements -- the scene, the characters, and the narrative structures -- to
express.
From the perspective of computer vision, movies provide a great arena with
a number of new challenges, \eg~substantially greater length,
richer presentation styles, and more complex temporal structures.
Recent studies~\cite{na2017read,rohrbach2015dataset,vicol2018moviegraphs,wang2018holistic,Nagrani17b,huang2018trailers} attempted to approach this problem from different
angles, only achieving limited progress.

%
Over the past decade, extensive studies have been devoted to video analytics.
A number of video-based tasks, \eg~action recognition~\cite{wang2016temporal,carreira2017quo} and event
classification~\cite{gan2015devnet}, have become active research topics.
However, methods devised for these tasks are not particularly suitable for
movie understanding. Specifically, for such tasks, visual features, which
can be a combination of various cues, are often sufficient for obtaining
good accuracies.
However, movies are essentially different. A movie is created to tell a story,
instead of demonstrating a scene or an event of a certain category.
To analyze movies effectively, we need new data, new perspectives, and thus
new approaches.

%
Recently, several datasets are constructed on movies, including
LSMDC~\cite{rohrbach2015dataset} and
MovieGraphs~\cite{vicol2018moviegraphs}.
These datasets, however, are limited in that they are small or have a
narrow focus on very short clips, \ie~those that last for a few seconds.
To facilitate the research in movie understanding, we need a new dataset
that is large and diverse, and more importantly allows high-level semantics
and temporal structures to be extracted and analyzed.
In this work, we construct a large dataset called
\emph{Movie Synopses Associations (MSA)} over $327$ movies.
This dataset not only provides a high-quality detailed synopsis for
each movie, but also associates individual paragraphs of the synopsis
with movie segments via manual annotation.
Here, each movie segment can last for several minutes
and capture a complete event.
These movie segments, combined with the associated synopsis paragraphs,
allow one to conduct analysis with a larger scope and at a higher semantic level.

%
Figure~\ref{fig:teaser} shows a movie segment and
the corresponding synopsis paragraph, where we have two important observations:
(1) The story is presented with a flow of events, governed by the underlying
narrative structures.
The sentences in the synopsis often follow a similar order.
(2) The characters and their interactions are the key elements of
the underlying story.
These two key aspects, namely the dynamic flow of events and the interaction
among characters, distinguish movies from those videos in
conventional tasks.

%
In this work, we develop a new framework for matching between
movie segments and synopsis paragraphs.
Rather than encoding them with feature vectors, we choose to use graphs
for representation, which provide a flexible way to capture middle-level
elements and the relationships among them.
Specifically, the framework integrates two key modules:
(1) \emph{Event flow module} for aligning the sequence of shots in a movie
segment, each showing a particular event, to the sequence of sentences
in a synopsis paragraph.
(2) \emph{Character interaction module} for capturing characters
and their behaviors (both actions and interactions) and associating them
with the corresponding descriptions.
Based on these two modules, the matching can then be done by solving
optimization problems formulated based on their respective
representations.

It is noteworthy that the use of graphs in movie representation has
been explored by previous works~\cite{vicol2018moviegraphs}.
However, our framework is distinguished in several aspects:
{1) It takes into account complicated temporal structures
and character interactions mined from data. 2) Our method does not
require node-to-node annotation when using graphs.

%
In summary, our contributions lie in three aspects:
(1) We construct a large dataset \emph{MSA} on $327$ movies, which provides
annotated associations between movie segments and synopsis paragraphs.
This dataset can effectively support the study on how movie segments are
associated with descriptions, which we believe is an important step towards
high-level movie understanding.
(2) We develop a graph-based framework that takes into account both
the flow of events and the interactions among characters.
Experiments show that this framework is effective, significantly improving the
retrieval accuracies compared to popular methods like
visual semantic embedding.
(3) We perform a study, which reveals the importance of high-level temporal structures
and character interactions in movie understanding. We wish that this study can motivate
future works to investigate how these aspects can be better leveraged.
\vspace{-2pt}

%
\section{Related Work}

%
\vspace{-2pt}
\paragraph{Datasets for Cross Modal Understanding.}
In recent years, with the increasing popularity of cross-modal understanding tasks,
\eg video retrieval by language, a large number of datasets have been proposed
~\cite{xu2016msr,anne2017localizing,rohrbach2015dataset,krishna2017dense,vicol2018moviegraphs,tapaswi2016movieqa,tapaswi2015book2movie,wang2016learning}.
\emph{ActivityNet Captions}~\cite{krishna2017dense} is 
a dataset with dense captions describing videos from \emph{ActivityNet} ~\cite{caba2015activitynet},
which can facilitate tasks such as video retrieval and temporal localization with language queries.
\emph{Large Scale Movie Description Challenge (LSMDC)} ~\cite{rohrbach2015dataset}
consists of short clips from movies described by natural language.
\emph{MovieQA}~\cite{tapaswi2016movieqa} is constructed for understanding
stories in movies by question answering. 
Some of the movies are provided plots with aligned movie clips.
\emph{MovieGraphs}~\cite{vicol2018moviegraphs} is established
for human-centric situation understanding with graph annotations.
But there are three problems for these datasets:
(1) most of them obtain dull descriptions from crowd-sourcing platforms,
(2) they simply describe short video clips lasting a few seconds,
which leads to a huge gap between proposed data and real-world data where the video is much longer and the description is much more complex.
(3) some of them are relatively smaller in terms of dataset size.
In order to explore the high-level
semantics and temporal structures in the data from real-world scenarios,
we build a new dataset with long segments cut
from movies and diverse descriptions from the synopses in IMDb\footnote{https://www.imdb.com}.
\vspace{-10pt}

%
\vspace{-2pt}
\paragraph{Feature-based Methods.}
To retrieve a video with natural language queries, 
the main challenge is the gap between two different modals. 
\emph{Visual Semantic Embedding} (VSE)~\cite{frome2013devise,faghri2017vse++},
a widely adopted approach in video retrieval~\cite{yu2017end,kordopatis2017near,Yu_2018_ECCV,dong2018dual,xu2019multilevel},
tries to tackle this problem by embedding multi-modal information into a common space.
JSF proposed in ~\cite{Yu_2018_ECCV} learns matching kernels
based on feature sequence fusion.
%
To retrieve video and localize clips, ~\cite{Shao_2018_ECCV} introduces a framework that
first perform paragraph level retrieval and then refine the features by sentence level clip localization. 
Feature-based approaches can not further improve retrieval performance because these
methods fail to capture the internal structures of 
video and language.
\vspace{-10pt}

\paragraph{Graph-based Methods.}
Graph-based methods~\cite{johnson2015image,lin2014visual,vicol2018moviegraphs},
which build semantic graphs from both language and video
and then formulate the retrieval task as a graph matching
problem~\cite{berg2005shape,zhou2012factorized,Zanfir_2018_CVPR},
is also widely used for cross-modal retrieval.
Method in \cite{johnson2015image} generates scene graph from language queries for image retrieval.
A graph matching algorithm is proposed by \cite{lin2014visual} for semantic search in the domain of autonomous driving. The graph matching problem is formulated 
as LP optimization with ground-truth alignment in optimization constraints.
MovieGraphs proposed in~\cite{vicol2018moviegraphs} uses graph
as semantic representation and integrates graph into potential functions
for training.
It's noteworthy that node-level annotations are required during
training.
In this work, we also use graph-based representations for both movies and synopses.
However, unlike previous works that depend on the costly node-level annotations,
our graph matching only needs ground-truth of paragraph-level alignment,
which makes it much more practical.

\begin{figure*}[t]
		\centering
		\includegraphics[width=0.8\linewidth]{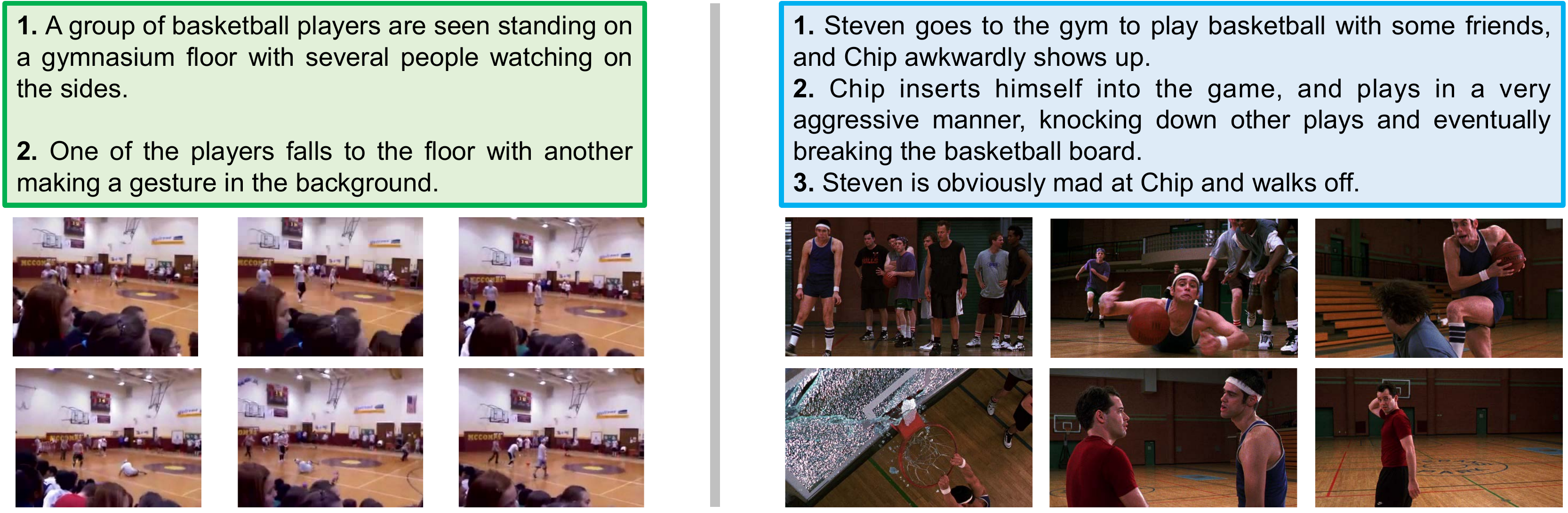}
		\caption{\small{Comparison between examples from ActivityNet Caption (left) and
		MSA (right). The durations are 12s and 220s respectively.}
		}
		\label{fig:data_compare}
			\vspace{-10pt}
\end{figure*}

\section{MSA Dataset}
\vspace{-5pt}
This section presents \emph{Movie Synopsis Association (MSA)},
a new dataset constructed upon $327$ movies.
Particularly, we choose a set of high-quality synopses from
IMDb,
\ie~those
with detailed descriptions of individual events, one for each movie.
Each synopsis here consists of tens of paragraphs,
each describing an event in the movie.

We also provide the associations between movie segments and synopsis
paragraphs through manual annotation. These associations constitute a solid
basis to support high-level semantic analysis.
We collected the associations following the procedure below.
(1) We provide the annotators with a complete overview of each movie,
including the character list, reviews, \etc,  to ensure they are familiar
with the movies.
(2) We carry out the annotation procedure in two stages, from coarse to fine.
At the first stage, each movie is divided into $64$ clips, each lasting
for around $2$ minutes.
For each synopsis paragraph, an annotator is asked to select a segment,
\ie~a subsequence of $N$ consecutive clips, that cover the
corresponding description.
At the second stage, annotators adjust the temporal boundaries of the resultant
segments to make them better aligned with the paragraphs.
This two-stage procedure leads to a collection of paragraph-segment pairs.
(3) We dispatch each paragraph to three annotators and only retain those
annotations with high consistency among them. Here, the consistency is measured
in terms of temporal IoU among the annotations.
Finally, we obtained $4,494$ highly consistent paragraph-segment pairs
(out of $5,725$ annotations of the original collection).

\begin{table}[]
	\caption{\small{Statistics of the \emph{MSA} dataset.}
	}
	\centering
	{\small
	\begin{tabular}{|l|c|c|c|c|}
		\hline
		& Train  & Val    & Test   & Total  \\ \hline
		\# Movies         & 249    & 28     & 50     & 327    \\
		\# Segments       & 3329   & 341    & 824    & 4494   \\
		\# Shots / seg.   & 96.4  & 89.8  & 76.9  & 92.3  \\
		Duration / seg.   & 427.4 & 469.6 & 332.8 & 413.3 \\
		\# Sents. / para. & 6.0   & 6.0   & 5.5   & 5.9   \\
		\# Words. / para. & 130.8 & 132.5 & 120.5 & 129.0 \\ \hline
	\end{tabular}
	}
	\vspace{-5pt}
	\label{tab:ds_stat}
\end{table}

\begin{table}[]
	\centering
	\caption{\small{Comparison between MSA dataset and MovieQA~\cite{tapaswi2016movieqa}.}}
	{\small
		\begin{tabular}{|r|cccc|}
			\hline
			\multicolumn{1}{|l|}{} & \multicolumn{1}{l}{\#movie} & \multicolumn{1}{l}{\#sent./movie} & \multicolumn{1}{l}{\#words/sent.} & \multicolumn{1}{l|}{dur. (s)} \\ \hline
			MovieQA               & 140                        & 35.2                             & 20.3                              & 202.7                          \\
			MSA                   & 327                        & 81.2                             & 21.8                              & 413.3                          \\ \hline
		\end{tabular}
	}
	\vspace{-10pt}
	\label{tab:mqa}
\end{table}

Table~\ref{tab:ds_stat} shows some basic statistics of the dataset.
This dataset is challenging:
(1) The duration of each movie segment is
over $400$ seconds on average, far longer than those in existing datasets
like LSMDC~\cite{rohrbach2015dataset}.
(2) The descriptions are rich with over $100$ words per paragraph.

Figure~\ref{fig:data_compare} compares ActivityNet Caption~\cite{krishna2017dense}
with the MSA dataset with examples.
We can see that the descriptions in MSA are generally much richer
and at a higher level, \eg~describing characters and events, instead of
simple actions.
MovieQA also contains description-clip pairs. Table~\ref{tab:mqa} compares
MovieQA with our MSA dataset. Note that the plot synopses from MovieQA are
obtained from Wikipedia while ours are from IMDb. 
Compared to synopses from Wikipedia, those from IMDb are written by movie fans and reviewed by others. They are longer and contain more details.


\section{Methodology}
\label{sec:method}

%

In this section,
we would present our framework for matching between movie segments and synopsis paragraphs.
Specifically, given a query paragraph $P$ from a synopsis,
we aim at retrieving its associated movie segment $Q$
out of a large pool of candidates.
This framework consists of two modules:
a \emph{Event Flow Module (EFM)}
to exploit the temporal structure of the event flows, and
a \emph{Character Interaction Module (CIM)} to leverage character
interactions.

As shown in Figure~\ref{fig:teaser},
given a query paragraph $P$ and a candidate movie segment $Q$,
each module yields a similarity score between $P$ and $Q$,
denoted as $\cS_{efm}(P,Q)$ and $\cS_{cim}(P,Q)$ respectively.
Then the overall matching score $\cS(P,Q)$ is defined to be their sum as
\begin{equation}
	\cS(P,Q) = \cS_{efm}(P,Q) + \cS_{cim}(P,Q),
\end{equation}

\begin{figure}[t]
	\centering
	\includegraphics[width=0.9\linewidth]{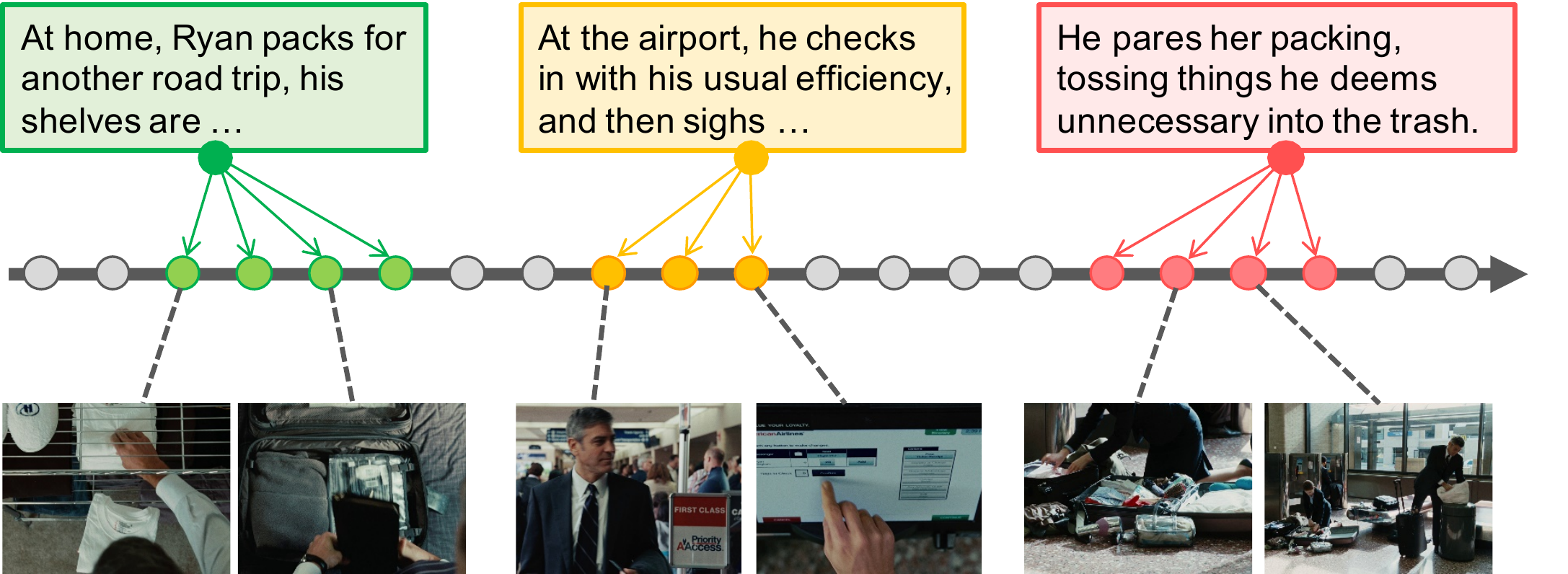}
	\caption{\small
		Sentences in a synopsis paragraph often follow a similar order
        as the situations in event presented in the movie segment. Therefore,
        they can be aligned temporally.
	}
	\label{fig:efm}
		\vspace{-10pt}
\end{figure}

In what follows,
Sec.~\ref{subsec:tam} and \ref{subsec:cim}
present the EFM and CIM modules respectively.
Sec.~\ref{subsec:em} introduces the training algorithm, where
both modules are jointly optimized.

\subsection{Event Flow Module}
%
\label{subsec:tam}

This module takes into account the temporal structures of event flows.
It is motivated by the observation that the sentences in a synopsis paragraph
tend to follow a similar order as that of situation in events (each captured by a  sequence of movie shots\footnote{A shot is a series of frames, that
runs for an uninterrupted period of time. Observing that frames within a shot
are highly redundant, we use shot as the unit instead of frames.}),
as shown in Figure~\ref{fig:efm}.
In particular, the alignment between the sentences and the movie shots
can be done based on the following principles:
(1) Each sentence can match multiple shots while
	a shot can be assigned to at most one sentence.
(2) The sentences and the movie shots follow the same order.
    The matching should not swap the order, \eg~associating a sentence
    that comes next to a preceding shot.


\vspace{3pt}
\noindent\textbf{Formulation.}
Suppose a paragraph $P$ is composed of a sequence of sentences $\{p_1, \dots, p_{M}\}$.
We obtain an embedding feature ${\vphi}_i \in \Rbb^D$ for each sentence $p_i$
using fully connected embedding networks.
Meanwhile, a movie segment $Q$ consists of a sequence of \emph{shots},
which can be extracted by a shot segmentation tool~\cite{sidiropoulos2011temporal}.
We derive a visual feature ${\vpsi}_i \in \Rbb^D$ for each shot $q_i$
with fully connected embedding networks.
Here we aim at assigning each sentence to a sub-sequence of shots,
which can be represented by a binary assignment
matrix $\mY \in \{0, 1\}^{N \times M}$,
where $y_{ij}=\mY(i,j)=1$ if the $i^{th}$ shot is attached to the $j^{th}$ sentence
and $0$ otherwise.
Given the assignment matrix $\mY$, the total matching score can be
expressed as
\begin{align} \label{eq:Stam}
\cS_{efm}=\sum_{i}\sum_{j}y_{ij} \vphi_j^T \vpsi_i=\mathrm{tr}(\mPhi \mPsi^T \mY ),
\end{align}
where
$\mPhi=[\vphi_1, \dots, \vphi_M]^T$ and  $\mPsi=[\vpsi_1, \dots, \vpsi_N]^T$
are the feature matrices for both domains.
Taking the alignment principles described above into account,
we can obtain the assignment $\mY$ by solving the following problem:
\begin{align} \label{eq:opt_tam}
\max_{\mY} \quad & \mathrm{tr}(\mPhi \mPsi^T \mY )\\
\textrm{s.t.} \quad & \mY \vone \preceq \vone,\\
&\cI(\vy_i) \leq \cI(\vy_{i+1}), \forall i \leq N-1.
\end{align}
Here, $\vy_i$ refers to the $i^{th}$ row of matrix $\mY$,
and $\cI(\cdot)$ denotes for the index of the first nonzero
element in a binary vector.
%
This is a bipartite graph matching problem which can be
efficiently solved by dynamic programming. 

\begin{figure}[t]
	\centering
	\includegraphics[width=0.9\linewidth]{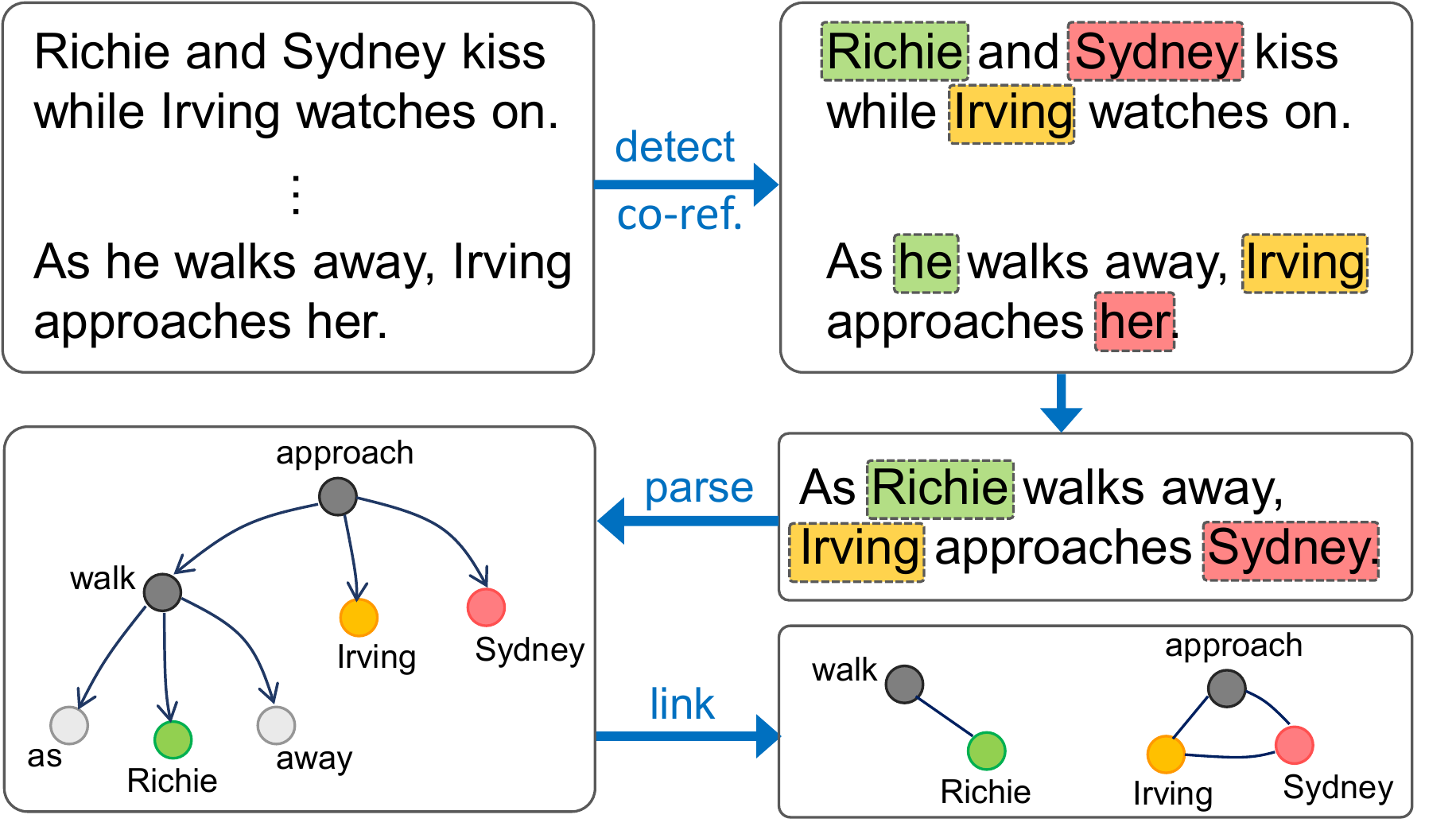}
	\caption{\small
		The procedure of constructing graphs from paragraph.
		At first, all the character names and pronouns are detected.
		Then each sentence is parsed to a dependency tree. Based on the
		tree structure, graphs are generated at rule-based linking stage.
	}
	\label{fig:gen_para_graph}
		\vspace{-10pt}
\end{figure}

\subsection{Character Interaction Module}
%
\label{subsec:cim}

As discussed earlier, the interactions among characters
play a significant role in movie storytelling. We also observe that
the character interactions are often described in synopsis.
To incorporate this aspect, we propose the
\emph{Character Interaction Module (CIM)} based on graph
representations derived from both the synopsis paragraphs and
the visual observations in the movie segments.

Specifically,
each paragraph and movie segment
are represented by graphs $\cG_p=(V_p, E_p)$ and
$\cG_q=(V_q, E_q)$ respectively.
The vertex sets $V_p$ and $V_q$ contain both character and action nodes.
The edge sets $E_p$ and $E_q$ capture both character-character
and character-action relations.
With these two graphs, the similarity between $P$ and $Q$ can be
computed by matching between $\cG_p$ and $\cG_q$. Below, we elaborate on
the matching procedure.

%
\vspace{-10pt}
\paragraph{Visual Graph from a Movie Segment.}
Firstly, we generate the character and action nodes:
(1) For character nodes, we utilize Faster-RCNN~\cite{girshick2015fast} implemented on~\cite{chen2019mmdetection} and 
pre-trained on~\cite{Huang_2018_CVPR,huang2018person}
to detect person instances in every shot.
(2) We attach each person instance with an action node,
which comes from a TSN~\cite{wang2016temporal} pretrained on \emph{AVA} dataset~\cite{gu2018ava}.
Secondly, we produce the edge sets by the following procedures:
(1) If a group of people appear in the same or adjacent shots,
we introduce an edge between every pair of them.
(2) We link each character node to its corresponding action node.

\vspace{-10pt}
%
\paragraph{Semantic Graphs from Sentences.}
For each paragraph, we construct a collections of sub-graphs from each sentence
based on dependency trees,
as illustrated in Figure~\ref{fig:gen_para_graph}.

The construction process consists of four major steps:
\textbf{(1) Name detection:}
We detect all the named entities (\eg, \emph{Jack})
using StanfordNer~\cite{finkel2005incorporating}.
Then we resort to CorefAnnotator~\cite{recasens_demarneffe_potts2013}
to link pronouns with named entities and substitute all pronouns with their corresponding names.
\textbf{(2) Character association:}
With the help of IMDb, we can retrieve a portrait for each named character
and thus obtain facial and body features using ResNet~\cite{he2016deep} pre-trained
on \emph{PIPA}~\cite{zhang2015beyond}.
This allows character nodes to be matched to the person instances detected in
the movie.
\textbf{(3) Sentence parsing:}
We use GoogleNLP API\footnote{https://cloud.google.com/natural-language/}
to obtain the dependency tree of a sentence.
Each node in the tree is labeled with a part-of-speech tagging.
\textbf{(4) Edge linking:}
Based on the dependency tree, we link each character name to its
parent verb. Meanwhile, if a group of character names share the same verb,
we introduce an edge between every pair of them. Note that
we only consider the verbs that stand for action. 
We first select $1000$ verbs with the highest frequency from the synopses corpus,
and then retain those corresponding to visually observable actions,
\eg ``run''. This results in a set of $353$ verbs.

It is worth noting that we generate a collection of sub-graphs from
paragraph instead of a connected graph. 
For convenience, we consider the collection of sub-graphs as a 
graph with notation $\cG_p$
although it can be further decomposed into multiple disjoint sub-graphs.
This is also what we do in our implementation.

\vspace{-10pt}
%
\paragraph{Matching Paragraph with Movie Segment.}
For graph $\cG_p$, let $V_p$ be its vertex set with $|{V_p}|=m=m_c+m_a$,
where $m_c$ is the number of character nodes and $m_a$ is that of action nodes.
Similarly, we have $\cG_q$ with $|{V_q}|=n=n_c+n_a$.

%
The target of graph matching is to establish a node-to-node assignment for the two input graphs while
taking the the pair-wise constraints, namely the edges, into account.

We define a binary vector $\vu \in \{0,1\}^{nm \times 1}$ as the indicator,
where $u_{ia}=1$ if $i \in V_q$ is assigned to $a \in V_p$.
To measure the similarity of nodes and edges from different graphs,
we establish the similarity matrix $\mK \in \Rbb^{nm \times nm}$,
where the diagonal elements represent node similarities
whereas the off-diagonal entries denote edge similarities.
Particularly, $\kappa_{ia;ia}=\mK(ia,ia)$ measures the similarity between $i^{th}$
node in $V_q$ and $a^{th}$ node in $V_p$.
$\kappa_{ia;jb}$ measures the similarity between two edges $(i,j)\in E_q$ and $(a,b)\in E_p$.
The nodes are represented as output features from networks. And the edge is represented by the concatenation of its nodes' features. The similarities in $\mK$ is computed by dot product between feature vectors.

Given the indicator $\vu$ and the similarity matrix $\mK$,
the similarity of two graphs can be derived as
\begin{equation}
\cS_{cim}(P,Q) = \sum_{i,a} u_{ia}\kappa_{ia;ia} + \
\sum_{\substack{i,j\\i\neq j}}\sum_{\substack{a,b\\a\neq b}}
u_{ia} u_{jb} \kappa_{ia;jb},
\end{equation}
where the first term models the similarity score between matched notes
$i\in V_q$ and $a\in V_p$.
The second term gives the bonus from matched edges between $(i,j)$ and $(a,b)$.

%
Based on the properties of nodes, certain constraints are enforced on $\vu$:
(1) The matching should be a one-to-one mapping.
For example, one node in a vertex set can only be
matched to at most one node in the other set.
%
(2) Nodes of different types cannot be matched together.
For example, a character node can not be assigned to
an action node.

The objective function, together with the constraints,
can be simply expressed in the following form:
\vspace{-5pt}
\begin{align} \label{eq:opt_cim}
\max_{\vu} \quad & \vu^T \mK \vu,\\
\textrm{s.t.} \quad & \textstyle \sum_{i}u_{ia} \leq 1 \quad\forall a,\\
& \textstyle \sum_{a}u_{ia} \leq 1 \quad\forall i, \\
& \textstyle \sum_{i \in V_q^c}u_{ia} = 0 \quad\forall a \in V_p^a, \\
& \textstyle \sum_{i \in V_q^a}u_{ia} = 0 \quad\forall a \in V_p^c.
\end{align}

Here $V_q^a$ denotes the vertex set containing only
action nodes in video with $|V_q^a|=n_a$ and $V_q^c$ for vertex
only containing cast nodes in video. The same for $V_p^a$ and $V_p^c$.

%

%
\vspace{-10pt}
\paragraph{Graph Pruning}
The problem itself is known as an NP-hard \emph{Quadratic Assignment Problem (QAP)}.
Solving it could be time consuming especially when the graph is large, which is normally the case for our video graph.
To ease the problem, we propose a graph pruning strategy
to reduce the graph size to an appropriate one that it can be solved
in an affordable time.
The strategy is described as follows:
%

\textbf{Seed Node Generation.}
	We first select the most important nodes as seed nodes. They are selected by the following two criteria:
	(a) The problem can be approximately solved by \emph{Kuhn--Munkres (KM)} algorithm~\cite{Kuhn1955} in polynomial time.
	The matched nodes can be selected as seed nodes.
	(b) The $k$ most similar nodes with each node from the query graph
	will be chosen as seed nodes.

\textbf{Selection Propagation.}
	Given the seed nodes, we extend the node selection by considering
	the nodes within $J^{th}$ degree connection of a seed node.
	We denote the seed nodes by another indicator vector $\vv \in \{0,1\}^{n\times 1}$,
	the adjacency matrix as $\cA$ of graph $\cG_q$, the nodes we select can be
	expressed as $\vv \gets \cA^J \vv $. The pruned graph is obtained by cropping the whole graph using selected nodes.
%

%
\subsection{Joint Optimization.}
\label{subsec:em}

The quality of the node features
would highly influence the result of matching.
It is necessary for us to finetune the parameters of the models in EFM and CIM for a better representations.
Since we do not have the ground truth alignment of $\mY$ in EFM
or $\vu$ in CIM, we can not directly update the model parameters in a
supervised manner.
Hence, we adopt an EM-like procedure to finetune the feature
representations and optimize matching objectives.
The overall loss of the whole framework is given below:
\vspace{-5pt}
\begin{equation}
\cL = \cL(\mY, \vtheta_{efm}, \vu, \vtheta_{cim})
\end{equation}

where $\vtheta_{efm}$ and $ \vtheta_{cim}$ denote model parameters for embedding networks
in EFM and CIM respectively. 

\vspace{-10pt}
\paragraph{E-Step.}
Using current model parameter values $\vtheta_{efm}^*$ and $ \vtheta_{cim}^*$,
we solve Eq.\ref{eq:opt_tam} by dynamic programming mentioned in
Sec.\ref{subsec:tam} and we obtain a sub-optimal value in Eq.\ref{eq:opt_cim}
by applying KM algorithm.
Here in our implementation, the time complexity of the KM algorithm
is $\cO(\tau^3)$ where $min(n,m)\leq \tau \leq max(n,m)$.

\vspace{-10pt}
\paragraph{M-Step.}
We update the model parameters in M-step with optimal solutions
 $\mY^*$ and $\vu^*$ obtained in E-step.
Particularly, given $\mY^*$ and $\vu^*$, we update model parameters by
\begin{small}
\begin{multline}
\begin{aligned}
\!\vtheta_{efm}^*, \vtheta_{cim}^*=\argmin_{\vtheta_{efm}, \vtheta_{cim}} \cL(\mY^*, \vtheta_{efm}, \vu^*, \vtheta_{cim}) \\
= \argmin_{\vtheta_{efm}, \vtheta_{cim}} \cL(S^*; \vtheta_{efm}, \vtheta_{cim})\quad\quad
\end{aligned}
\end{multline}
\end{small}
where $\cL(S; \vtheta)$ is the pair-wise ranking loss with margin $\alpha$ shown below:
\vspace{-5pt}
\begin{multline}
\label{eq:loss}
\cL(S; \vtheta) = \sum_{i}\sum_{j\neq i} max(0, S(Q_j, P_i) - S(Q_i, P_i) + \alpha)\\
+ \sum_{i}\sum_{j\neq i} max(0, S(Q_i, P_j) - S(Q_i, P_i) + \alpha)
\end{multline}
\vspace{-10pt}

\begin{table*}[]
	\centering
	\caption{\small The overall performance of video
	retrieval on MSA dataset under both cross-movie and within-movie settings. 
	Here, \emph{appr.} refers to 
	appearance node, \emph{cast} stands for character node and
	\emph{action} denotes action node.}
\begin{tabular}{lll|cccc|cccc}
	\hline
	&                                    &                   & \multicolumn{4}{c|}{Cross-movie}                              & \multicolumn{4}{c}{Within-movie}                               \\ \hline
	& \multicolumn{1}{l|}{Method}        & Nodes             & R@1            & R@5            & R@10           & MedR       & R@1            & R@5            & R@7            & Avg. MedR    \\ \hline
	\multicolumn{1}{l|}{1} & \multicolumn{1}{l|}{Random}        & -                 & 0.12          & 0.61           & 1.21           & 412.5      & 6.07           & 28.88          & 38.35          & 8.74         \\ \hline
	\multicolumn{1}{l|}{2} & \multicolumn{1}{l|}{JSF}           & appr.             & 3.52           & 12.62          & 20.02          & 55         & 19.42          & 56.07          & 66.51          & 3.86         \\ \hline
	\multicolumn{1}{l|}{3} & \multicolumn{1}{l|}{VSE}           & appr.             & 4.49           & 15.41          & 24.51          & 39.5       & 21.36          & 60.07          & 69.42          & 3.62         \\ \hline
	\multicolumn{1}{l|}{4} & \multicolumn{1}{l|}{VSE}           & appr.+action      & 5.34           & 15.78          & 24.64          & 42.5       & 21.85          & 61.41         & 69.66          & 3.47         \\ \hline
	\multicolumn{1}{l|}{5} & \multicolumn{1}{l|}{VSE}           & appr.+action+cast & 19.05          & 48.67          & 60.92          & 6          & 26.70          & 65.90          & 72.94          & 3.03         \\ \hline\hline
	\multicolumn{1}{l|}{6} & \multicolumn{1}{l|}{Ours(EFM)}     & appr.             & 6.80           & 20.15          & 28.40         & 36         & 27.67          & 63.59          & 71.97          & 2.92         \\ \hline
	\multicolumn{1}{l|}{7} & \multicolumn{1}{l|}{Ours(EFM)}     & appr.+action+cast & 21.12          & 48.67          & 61.04          & 6          & 30.58 & 66.14          & 73.42          & 2.70 \\ \hline
	\multicolumn{1}{l|}{8} & \multicolumn{1}{l|}{Ours(EFM+CIM)} & appr.+action+cast & \textbf{24.15} & \textbf{53.28} & \textbf{66.75} & \textbf{4.5} & \textbf{31.92}          & \textbf{67.96} & \textbf{74.76} & \textbf{2.50}         \\ \hline
\end{tabular}
\vspace{-10pt}
	\label{tab:overall}
\end{table*}

\section{Experiments}
%
We conduct experiments of movie-synopsis retrieval on MSA dataset.
Specifically, search a movie segment from candidate pool
 given a synopsis paragraph as query.

%
\subsection{Experiment Setup}
\paragraph{Dataset.}
The MSA dataset is randomly split into \emph{train, val, test} subsets
with $3329, 341, 824$ samples respectively.
Note that there are no overlap movies among subsets. 
The statistic of the subsets is shown in Table~\ref{tab:ds_stat}.

There are two settings to measure the performance,
namely, \textbf{cross-movie} and \textbf{within-movie}.
The cross-movie setting considers the whole test set as the 
candidate pool for each query whereas the within-movie setting only
takes the segments from the same queried movie to be the candidates.
\vspace{-25pt}
\paragraph{Evaluation Metrics.}
%
%
To evaluate the performance, we adopt the commonly used metrics:
(1) \textbf{Recall@K}: the fraction of GT videos that 
have been ranked in top K;
(2) \textbf{MedR}: the median rank of GT videos.
(3) \textbf{Avg. MedR}: Average MedR, this is only for within-movie setting.

\vspace{-15pt}
%
\paragraph{Implementation Details.}
In EFM, Word2Vec~\cite{mikolov2013distributed} embedding is used as sentence representation.
The Word2Vec model is finetuned on MSA corpus, \ie, synopses and subtitles.
The shot feature consists of two parts:
1) visual features extracted from $pool5$ layer of ResNet-101~\cite{he2016deep}.
2) its subtitle's Word2Vec embedding.
In CIM, we adopt ResNet-50 pre-trained on \emph{PIPA}~\cite{zhang2015beyond}
to extract the face and body feature for a detected person instance or a cast portrait. The action features in videos come from TSN~\cite{wang2016temporal}
pre-trained on AVA~\cite{gu2018ava} and action verbs 
are represented by Word2Vec embeddings. 
We train all the embedding networks using SGD with learning rate $0.001$. 
The batch size is set to 16 and the margin $\alpha$ in pair-wise ranking
loss is set to $0.2$.

\subsection{Overall Results}
We adopt VSE as the base models and
previous method JSF~\cite{Yu_2018_ECCV} is also used for comparison.
Also for comparison, we gradually add three kinds of features, namely, \emph{ appearance,
cast} and \emph{action} as nodes to baseline method.
Particularly, appearance node denotes the sentence embeddings or shot
features.
%
For VSE, the features of movie shots and sentences are further transformed with
two-layer MLPs. We then obtain the features of segments and paragraphs by taking the average of the shot and sentence features. During matching, the segment/paragraph similarities are computed with cosine similarity. We use the same loss as shown in
Eq.~\ref{eq:loss}.
Matching scores from different nodes are fused by weighted sum.
The weights are obtained by observing the performance of single node on val set.
Here, for cross-movie setting, weights are simply set as $0.3, 1.0$ and $0.1$ for 
appearance, cast and action respectively. 
For within-movie setting, weights are $0.3, 0.3$ and $0.1$.
Table~\ref{tab:overall} shows the overall results of video retrieval
on MSA.
%

\vspace{-10pt}
\paragraph{Analysis on Overall Results.}
From the results shown in Table~\ref{tab:overall}, by
comparing different methods, we observe that:
	
(1) Both VSE and JSF outperform random guess by a large margin.
	The performance of JSF does not exceed that of VSE because
	the learned kernels in JSF fail to capture the matching pattern between 
	paragraphs and long videos, when the concepts in paragraphs
	are complicated and lengths of videos vary a lot.

(2) Our method with EFM and CIM outperforms the conventional
	methods that only fuse features under both cross-movie and within-movie
	settings. 
	Particularly, Recall@1 under cross-movie setting is raised from $19.05\%$
	to $24.15\%$ ($5.10\%$ absolute and $27\%$ relative improvement) and
	each recall under within-movie setting improves over $1.5\%$.

\vspace{-10pt}
\paragraph{Analysis on EFM and CIM.}
Also shown in Table~\ref{tab:overall},
the results of rows 3,6 demonstrate that
the proposed EFM improves the performance on most
of the metrics. We can see from the table that EFM
works better especially under within-movie setting 
($6.31\%$ increment on Recall@1).
It is because that encoded story and 
narrative structure in EFM is the key to distinguish segments 
from the same movie. 

Meanwhile, results from rows 7-8 
prove the effectiveness of using character interaction graph,
especially under cross-movie setting. 
The CIM does not bring consistent performance gain
under within-movie setting compared to EFM. 
The reason is that segments from the same
movie share a group of characters and their interactions are 
also similar. This is also illustrated in the right part of rows 4-5.

\begin{figure*}[!t]
	\centering
	\includegraphics[width=0.9\linewidth]{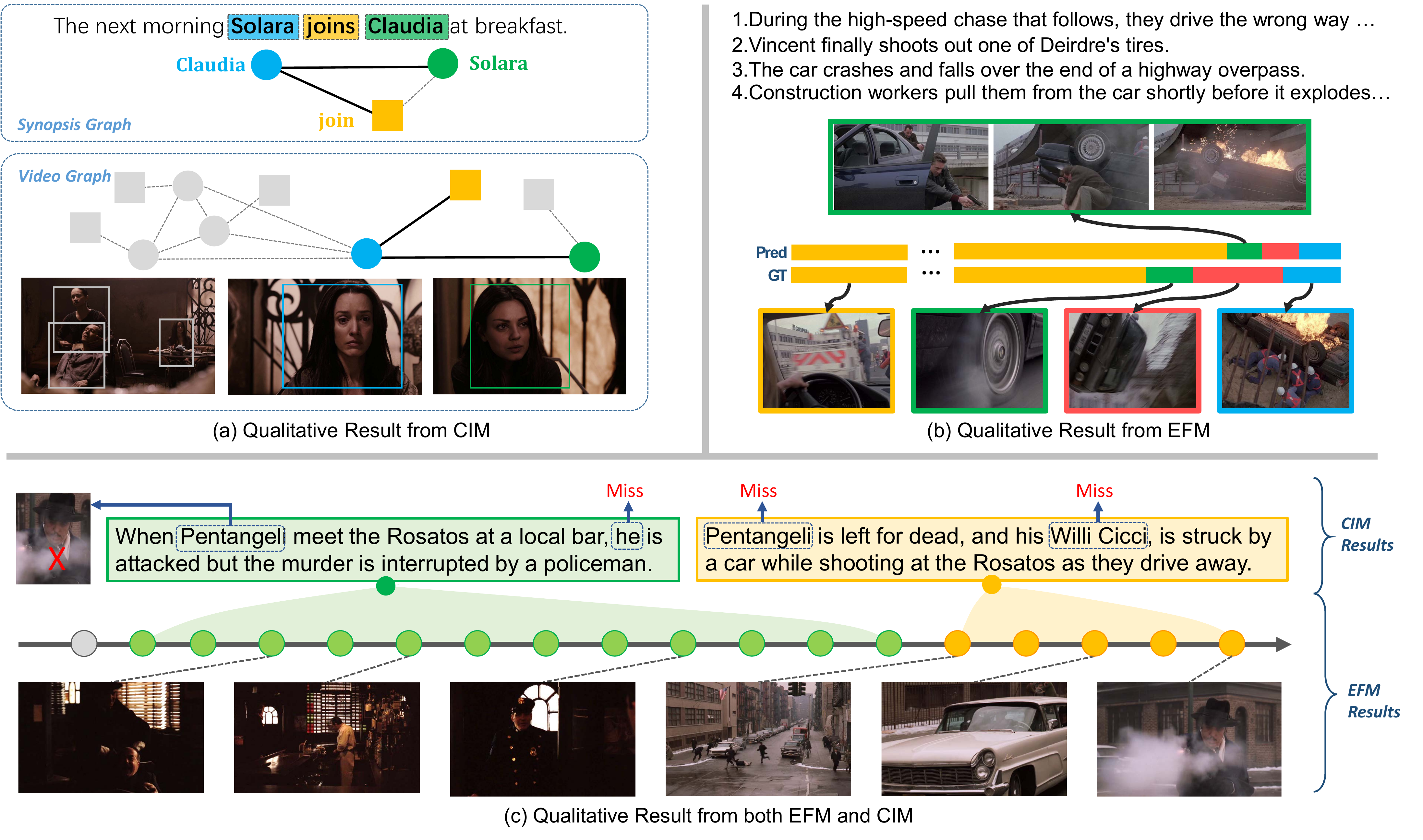}
	\caption{\small
		Qualitative results of EFM and CIM modules.
		(a) shows a success case of CIM; (b) presents a failure
		case of EFM; (c) shows an example that EFM succeeds but
		CIM fails.
	}
	\label{fig:qualitative}
	\vspace{-10pt}
\end{figure*}

\vspace{-3pt}
\subsection{Ablation Studies}
\vspace{-5pt}
We present ablation studies on different hyper parameters.
Unless stated, experiments are conducted under cross-movie setting.

\begin{table}[]
	\centering
	\caption{Influence of different choices of $N$ for updating scores in CIM.
	The first row is the result before updating.}
	\begin{tabular}{l|cccc}
		\hline
		& R@1   & R@5   & R@10  & MedR \\ \hline
		previous stage & 21.12 & 48.67 & 61.04 & 6    \\ \hline
		$N=15$          & \textbf{24.15} & \textbf{53.28} & \textbf{66.75} & \textbf{4.5}    \\ \hline
		$N=40$          & 23.91 & 51.94 & 63.71 & 5    \\ \hline
		$N=60$           & 23.42 & 51.46 & 63.11 & 5    \\ \hline
		$N=80$          & 23.42 & 51.46 & 62.86 & 5    \\ \hline
	\end{tabular}
	\label{tab:choose_n}
\end{table}

\vspace{-12pt}
\paragraph{Choices of $N$ in CIM.}
As mentioned before, at inference stage, we need to obtain
score in CIM by solving the optimization problem in Eq.\ref{eq:opt_cim}. 
It takes 2 seconds to solve one matching on average.
Under the cross-movie setting, we need to solve
these problems for $824^2$ times (the number of test samples is $824$), which sums up
to more than a week.
To save time, we only update the score of
candidates that rank top $N$ in previous stage, \eg, VSE with score fusion.

Table~\ref{tab:choose_n} shows the influence on different choices of $N$.
Note that we take the score in the first row to filter out
a candidate list for updating.
We see that 
from $N=15$ to $N=40$, the performance drops  while
remains steady when $N$ increases from $40$ to $80$.
All the results still outperform the baseline in the first row.
The performance drop comes from the increasing outliers when $N$ increases.
Therefore, decrease $N$ can not only improve inference efficiency
but also decrease the number of distractors in candidate pool.

\begin{table}[]
	\centering
	\caption{\small Comparison between the performance of using
	only visual feature and that of using both visual and subtitle features
	as shot representation. The input node is \emph{appr.}}
	\begin{tabular}{l|cccc}
		\hline
		& R@1  & R@5   & R@10  & \multicolumn{1}{l}{MedR} \\ \hline
		visual only     & 4.25 & 13.84 & 19.66 & 56                     \\ \hline
		visual + subtt. & \textbf{4.49} & \textbf{15.41} & \textbf{24.51} & \textbf{39.5}                     \\ \hline
	\end{tabular}
\vspace{-10pt}
	\label{tab:subtitle}
\end{table}

\vspace{-10pt}
\paragraph{Influence of using subtitle feature.}
Recall that we use both the visual and subtitle feature
as the representation of a shot by
observing that sometimes the narrators
tend to summarize important dialogues in synopses.
We conduct ablation study on the effectiveness of subtitle
feature shown in Table~\ref{tab:subtitle}.
The experiments are based on appearance nodes only.
The results show that subtitle are complementary
to visual information. 

\begin{table}[]
	\centering
	\caption{\small Comparison of  
	different graph pruning parameters.}
	\begin{tabular}{l|cccc}
		\hline
		& R@1   & R@5   & R@10  & MedR \\ \hline
		$J=1$ & 23.30 & 53.03 & 66.14 & 5    \\ \hline
		$J=2$ & \textbf{24.15} & \textbf{53.28} & \textbf{66.75} & \textbf{4.5}    \\ \hline
		$J=3$ & 24.03 & 53.16 & 66.63 & 5    \\ \hline
	\end{tabular}
\vspace{-10pt}
	\label{tab:prune}
\end{table}
\vspace{-10pt}
\paragraph{Graph Pruning Parameters.}
To raise inference efficiency, we perform graph pruning in CIM.
We set $k=2$ to select seed and $J=2$ to spread
selection (recall Sec.~\ref{subsec:cim}). As $k$ and $J$ are complementary for controlling the size
of pruned graph, we only conduct studies on different values of $J$. 
The results are shown in Table~\ref{tab:prune}.
It demonstrates that $J=2$ is enough for pruning a graph and increase
$J$ may introduce more noise.

\subsection{Qualitative Results}
\vspace{-5pt}
We present qualitative results on both EFM and CIM modules
to further explore their effectiveness.

Figure~\ref{fig:qualitative} (a) shows a positive result that
the characters and actions in the sentence are accurately matched.
The right matching is obtained with 
the help of character-character and character-action relations.

Figure~\ref{fig:qualitative} (c) shows a case that EFM
successfully assigns each sentence to the corresponding
shots while CIM fails to assign the characters.
In particular, ``Pentangeli'' is assigned to a wrong person
instance while the other three names match nothing.
The reason is that the person instances from movie segment are in
poor quality due to dim light, occlusion or large motion
expect the one appearing at the end of the segment.

Figure~\ref{fig:qualitative} (b) shows a failure case of 
EFM where the second sentence is completely miss-aligned.
As shown in the upper part of the figure, 
this is possible because the shots belong to the third
sentence contain some content of ``shoot'' and ``tire''
which mislead the model. We also observe that this
case is challenging because the shots look similar
to each other due to no transition of scene.

From the above observations and analysis on more such cases,
we come to the following empirical conclusions:
(1) Edge constraints are important for alignments.
(2) The quality of nodes matters. If nodes are in poor quality,
 the edge constraints will take no effect.
(3) Discriminative shot appearance, together with our proposed
EFM, is helpful for temporal alignment.



\section{Conclusion}
\vspace{-5pt}
In this paper, we propose a new framework for matching between movie
segments and synopsis paragraphs. The proposed framework
integrates a \emph{Event Flow Module} to capture
the \emph{narrative structures} of movies and a
\emph{Character Interaction Module} to model
character interactions using graph-based formulation.
To facilitate research for movie-synopsis matching, we construct a dataset
 called Movie Synopses Associations (\emph{MSA}).
Experimental results show the effectiveness of the proposed
modules. Our framework outperforms conventional feature-based
methods and improves the matching accuracy consistently on all metrics.
Both quantitative and qualitative studies demonstrate that 
our method can capture rich temporal structures and 
diverse interactions among characters.

\section{Acknowledgment}
This work is partially supported by the Collaborative Research grant from SenseTime Group (CUHK Agreement No. TS1610626 \& No. TS1712093), and the General Research Fund (GRF) of Hong Kong (No. 14236516 \& No. 14203518).

{\small
	\bibliographystyle{ieee_fullname}
	\bibliography{egbib}
}

\end{document}